# Machine Learning for Bioclimatic Modelling

Maumita Bhattacharya

School of Computing & Mathematics

Charles Sturt University

NSW, Australia – 2640

*Abstract*—**Many machine learning (ML) approaches are widely used to generate bioclimatic models for prediction of geographic range of organism as a function of climate. Applications such as prediction of range shift in organism, range of invasive species influenced by climate change are important parameters in understanding the impact of climate change. However, success of machine learning-based approaches depends on a number of factors. While it can be safely said that no particular ML technique can be effective in all applications and success of a technique is predominantly dependent on the application or the type of the problem, it is useful to understand their behavior to ensure informed choice of techniques. This paper presents a comprehensive review of machine learning-based bioclimatic model generation and analyses the factors influencing success of such models. Considering the wide use of statistical techniques, in our discussion we also include conventional statistical techniques used in bioclimatic modelling.**

*Keywords*—*Machine Learning; Bioclimatic Modelling; Geographic Range; Artificial Neural Network; Evolutionary Algorithm*

## I. INTRODUCTION

Understanding species' geographic range has become all the more important with concerns over global climatic changes and possible consequential range shifts, spread of invasive species and impact on endangered species. The key methods used to study geographic range are bioclimatic models, alternatively known as envelope models (Kadmon et al., 2003), climate response surface models (Huntley, 1995), ecological niche models (Peterson & Vieglais, 2001) or species distribution models (Loiselle et al., 2003). Predictive ability lies at the core of such methods as it is the ultimate goal of ecology (Peters, 1991).

Machine Learning (ML) as a research discipline has roots in Artificial Intelligence and Statistics and the ML techniques focus on extracting knowledge from datasets (Mitchell, 1997). This knowledge is represented in the form of a model which provides description of the given data and allows predictions for new data. This predictive ability makes ML a worthy candidate for bioclimatic modelling. Many ML algorithms are showing promising results in bioclimatic modelling including modelling and prediction of species distribution (Elith et al., 2006).

There are diverse applications of ML algorithms in ecology. They range from experimenting bio-geographical, ecological, and also evolutionary hypotheses to modelling species distributions for conservation, management and future planning (e.g., Fielding, 1999; Recknagel, 2001, 2003; Cushing and

Wilson, 2005; Ferrier and Guisan, 2006; Park and Chon, 2007). Under the broad umbrella of Eco-informatics (Green et al., 2005) machine learning (ML) is a fast growing area which is concerned with finding patterns in complex, often nonlinear and noisy data and generating predictive models of relatively high accuracy. The increase in use of the ML techniques in ecological modelling in recent years is justified by the fact that this ability to produce predictive models of high accuracy does not involve the restrictive assumptions required by conventional, parametric approaches (Guisan and Zimmermann, 2000; Peterson and Vieglais, 2001; Olden and Jackson, 2002a; Elith et al., 2006).

It may be noted that there is no universally best ML method; choice of a particular method or a combination of such methods is largely dependent on the particular application and requires human intervention to decide about the suitability of a method. However, concrete understanding of their behavior while applied to bioclimatic modelling can assist selection of appropriate ML technique for specific bioclimatic modelling applications.

In this paper we present a concise review of application of machine learning approaches to bioclimatic modelling and attempt to identify the factors that influence success or failure of such applications. In our discussion we have also included popular applications of statistical techniques to bioclimatic modelling.

The rest of the paper is organized as follows: Section II provides an overview of the Machine Learning and statistical methods commonly used in bioclimatic modelling and their applications to bioclimatic modelling; Section III presents an investigation on factors which influence success of such applications; finally in Section IV, we present some concluding remarks.

## II. ML & STATISTICAL TECHNIQUES AND THEIR APPLICATION TO BIOCLIMATIC MODELLING

The inference mechanisms employed by Machine Learning (ML) techniques involve drawing conclusions from a set of examples. Supervised learning is one of the key ML inference mechanisms and is of particular interest in prediction of geographic ranges. In supervised learning the information about the problem being modeled is presented by datasets comprising of input and desired output pairs (Mitchell, 1997). The ML inference mechanism extracts knowledge representation from these examples to predict outputs for new inputs. The ML inference mechanism is depicted in Fig. 1.

The relatively more popular bioclimatic modelling applications of statistical and machine learning techniques and features of the relevant techniques are discussed next.

### A. Statistical Approaches

#### 1) Generalised Linear Model (GLM)

Generalised linear models (GLM) (McCullagh and Nelder, 1989) are probably the most commonly used statistical method in the bioclimatic modelling community and have proven ability to predict current species distribution (Bakkenes et al., 2002).

Generalised linear model (GLM) is a flexible generalization of regular linear regression. In GLM the response variable is normally modeled as a linear function of the independent variables. The degree of the variance of each measurement is a function of its predicted value.

Logistic regression analysis has been widely used in many disciplines including medical, social and biological sciences (Hosmer and Lemeshow, 2000). Its bioclimatic modelling application is relatively straightforward where a binary response variable is regressed against a set of climate variables as independent variables.

#### 2) Generalised Additive Model (GAM)

Considering the limitations of Generalised Linear Models in capturing complex response curves, application of Generalised Additive Models is being proposed for species suitability modelling (Austin and Meyers, 1996; Seone et al., 2004; Austin, 2007).

The Generalised Additive Model (GAM) blends the properties of the Generalised Linear Models and Additive models (Friedman et al., 1981). GAM is based on non-parametric regression and unlike GLM does not impose the assumption that the data supports a particular functional form (normally linear) (Hastie and Tibshirani, 1990). Here the response variable is the additive combination of the independent variables' functions. However, transparency and interpretability are compromised to accommodate this greater flexibility.

GAM can be used to estimate a non-constant species' response function, where the function depends on the location of the independent variables in the environmental space.

#### 3) Climate Envelope Techniques

There are a number of specialized statistics-based tools developed for the purpose of bioclimatic modelling. Climate envelope techniques such as ANUCLAM, BIOCLIM, DOMAIN, FEM and HABITAT are popular and specialized bioclimatic modelling tools and thus deserve mention here. These tools usually fit a minimal envelope in a multidimensional climate space. Also, they use presence-only data instead of presence-absence data. This is highly beneficial as many data sets contain presence-only data.

Other statistical methods gaining popularity includes the Multivariate Adaptive Regression Splines (Elith et al., 2007).

### B. Machine Learning Approaches

#### 1) Evolutionary Algorithms (EA)

Evolutionary Algorithms are basically stochastic and iterative optimisation techniques with metaphor in natural evolution and biological sexual reproduction (Holland, 1975; Goldberg, 1989). Over the years several algorithms have been developed which fall in this category; some of the more popular ones being Genetic Algorithm, Evolutionary Programming, Genetic Programming, Evolution Strategy, Differential Evolution and so on. The most popular and extensive application of Evolutionary Algorithm and more specifically Genetic Algorithm (GA) to bioclimatic modelling has been through the software Genetic Algorithm for Rule-set Production (GARP) (Anderson et al., 2003; Peterson et al., 2001, 2002). Here, we shall restrict our discussion on application of Evolutionary Algorithm to bioclimatic modelling primarily to GARP.

Genetic Algorithm for Rule Set Production (Stockwell and Peters, 1999) is a specialised software based on Genetic Algorithm (Mitchell, 1999) for ecological niche modelling. The GARP model is represented by a set of mathematical rules based on environmental conditions. Each set of rules is an individual in the GA population. These rule sets are evolved through GA iterations. The model predicts presence of a species if all rules are satisfied for a specific environmental condition. The four sets of rules which are possible are: atomic, logistic regression, bio-climatic envelope and negated bio-climatic envelope (Lorena et al., 2011).

GARP is essentially a non-deterministic approach that produces Boolean responses (presence/absence) for each environmental condition. As in case of the climate envelope techniques, GARP also does not require presence/absence data and can handle presence-only data. However, as the "learning" in GARP is based on optimisation of a combination of several types of models and not of one particular type of model, ecological interpretability may be difficult.

Examples of applications of GARP for bioclimatic modelling include: the habitat suitability modelling of threatened species (Anderson and Martı́nez-Meyer, 2004) and that of invasive species (Peterson and Vieglais, 2001; Peterson, 2003; Drake and Lodge, 2006), and the geography of disease transmission (Peterson, 2001).

Other applications of Ganetic Algorithm to ecological modelling include: modelling of the distribution of cutthroat and rainbow trout as a function of stream habitat characteristics in the Pacific Northwest of the USA (D'Angelo et al., 1995) and modelling of plant species distributions as a function of both climate and land use variables (Termansen et al., 2006). McKay (2001) used Genetic Programming (GP) to develop spatial models for marsupial density. Chen et al. (2000) used GP to analyse fish stock-recruitment relationship, and Muttil and Lee (2005) used this technique to model nuisance algal blooms in coastal ecosystems. Newer approaches to use Evolutionary Algorithms for ecological niche modelling are being proposed such as the WhyWhere algorithm advocated by Stockwell (2006). EC has also been applied in conservation planning for biodiversity (Sarkar et al., 2006).

Fig. 1. Steps Involved In The Machine Learning Inference Process

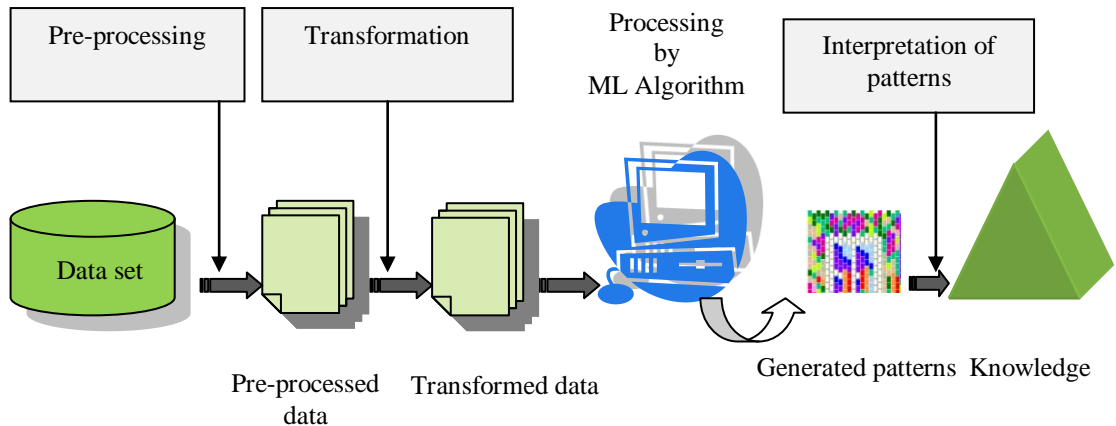

### 2) Artificial Neural Network (ANN)

A relatively later introduction to species distribution modelling is that of the Artificial Neural Network (ANN) (Manel et al., 1999; Olden et al., 2002b; Pearson et al., 2002; Thuiller, 2003).

Artificial Neural Networks are computational techniques with metaphor in the structure, processing mechanism and learning ability of the brain (Haykin, 1998). The processing units in ANN simulate biological neurons and are known as nodes.

These artificial neurons or nodes are organised in one or more layers. Simulating the biological synapses, each node is connected to one or more nodes through weighted connections. These weights are adjusted to acquire and store knowledge about data. There are many algorithms available to train the ANN.

Some of the noteworthy applications of ANN are as follows: species distribution modelling (Mastrorillo et al., 1997; O´zesmi and O´zesmi, 1999), species diversity modelling (Gue´gan et al., 1998; Brosse et al. 2001; Olden et al. 2006b), community composition modelling (Olden et al. 2006a), aquatic primary and secondary production modelling (Scardi and Harding 1999; McKenna 2005), species classification in appropriate taxonomic groups using multi-locus genotypes (Cornuet et al., 1996), modelling of wildlife damage to farmlands (Spitz and Lek , 1999), assessment of potential impacts of climate change on distribution of tree species in Europe (Thuiller, 2003), invasive species modelling (Vander Zanden et al. 2004), and pest management (Worner and Gevrey, 2006). Please see Olden et al. (2008) for further details.

The main advantages of ANNs are that they are robust, perform well with noisy data and can represent both linear and non-linear functions of different forms and complexity levels. Their ability to handle non-linear responses to environmental variables is an advantage.

However, they are less transparent and difficult to interpret. Inability to identify the relative importance and effect of the individual environmental variables is a limitation (Thuiller, 2003).

### 3) Decision Trees (DT)

Decision Trees have also found numerous applications in bioclimatic modelling. Decision Trees represent the knowledge extracted from data in a recursive, hierarchical structure comprising of nodes and branches (Quinlan, 1986). Each internal node represents an input variable or attribute. They are associated with a test or decision rule relevant to data classification. Each leaf node represents a classification or a decision i.e. the value of the target variable conditional to the value of the input variables represented by the root to leaf path. Predictions derived from a Decision Tree generally involve determination of a series of 'if-then-else' conditions (Breiman et al., 1984).

The two main types of Decision Trees used for predictions are: Classification Tree analysis and Regression Tree Analysis. The term Classification and Regression Tree (CART) analysis is the umbrella term used to refer to both Classification Tree analysis and Regression Tree analysis (Breiman et al., 1984).

Some of the relevant and relatively recent applications of Decision Trees are as follows: habitat models for tortoise species (Anderson et al. 2000), and endangered crayfishes (Usio, 2007); quantification of the relationship between frequency and severity of forest fires and landscape structure by Rollins et al. (2004); prediction of fish species invasions in the Laurentian Great Lakes by Mercado-Silva et al. (2006); species distribution modelling of bottlenose dolphin (Torres et al., 2003);development of models to assess the vulnerability of the landscape to tsunami damage (Iverson and Prasad, 2007). Olden et al. (2008) provides a more complete list.

The obvious advantage of the Decision Trees is that the ecological interpretability of the results derived from them is simple. Also there are no assumed functional relationships between the environmental variables and species suitability

TABLE I.     COMPARISON OF SOME OF THE RELEVANT CHARACTERISTICS OF ML TECHNIQUES

| Characteristic | GLM | DT | ANN | EA |
|---|---|---|---|---|
| *Mixed data handling ability* | Low | High | Low | Moderate |
| *Outlier handling ability* | Low | Moderate | Moderate | Moderate |
| *Non-linear relationship modelling* | Low | Moderate | High | High |
| *Transparency of modelling process* | High | Moderate | Low | Low |
| *Predictive ability* | Low | Moderate | High | High |

(De'Ath and Fabricius, 2000; Roguet et al., 2001; Vayssieres et al., 2000).

Despite their ease of interpretability, Decision Trees may suffer from over-fitting (Breiman et al., 1984; Thuiller, 2003).

Some relevant characteristics of different ML techniques are depicted in Table 1. Also see Olden et al. (2008).

## III. FACTORS INFLUENCING SUCCESS OF ML APPROACHES TO BIOCLIMATIC MODELLING

While it is not that straightforward to identify the causes of success or failure of applications of the Machine Learning techniques to bioclimatic modeling, in this section we attempt to outline some of the factors which may impact their performance broadly. However, this is not to undermine the fact that success or failure of any machine learning application is predominantly dependent on the specific application.

### A. Very large data sets

Data sets with hundreds of fields and tables and millions of records are commonplace and may pose challenge to the ML processors. However, enhanced algorithms, effective sampling, approximation and massively parallel processing offer solution to this problem.

### B. High dimensionality

Many bioclimatic modeling problems may require a large number of attributes to define the problem. Machine learning algorithms struggle when they are to deal with not just large data sets with millions of records, but with a large number of fields or attributes, increasing the dimensionality of the problem. A high dimensional data set pose challenges by increasing the search space for model induction. This also increases the chances of the ML algorithm finding invalid patterns. Solution to this problem includes reducing dimensionality and using prior knowledge to identify irrelevant attributes.

### C. Over-fitting

Over-fitting occurs when the algorithm can model not only the valid patterns in the data but also any noise specific to the data set. This leads to poor performance as it can exaggerate minor fluctuations in the data. Decision Tress and also some of the Artificial Neural Networks may suffer from over-fitting.

Cross-validation and regularization are some of the possible solutions to this problem.

### D. Dynamic environment

Rapidly changing or dynamic data makes it hard to discover patterns as previously discovered patterns may become invalid. Values of the defining variables may become unstable. Incremental methods that are capable of updating the patterns and identifying the patterns of changes hold the solution.

### E. Noisy and missing data

This problem is not uncommon in ecological data sets. Data smoothing techniques may be used for noisy data. Statistical strategies to identify hidden variables and dependencies may also be used.

### F. Complex dependencies among attributes

The traditional Machine Learning techniques are not necessarily geared to handle complex dependencies among the attributes. Techniques which are capable of deriving dependencies between variables have also been experimented in the context of data mining (Dzeroski, 1996; Djoko et al., 1995).

### G. Interpretability of the generated patterns

Ecological interpretability of the generated patterns is a major issue in many of the ML applications to bioclimatic modeling. Applications of Evolutionary Algorithm and Artificial Neural Networks may suffer from poor interpretability. Decision Trees on the other hand scores high in terms of interpretability.

Other influencing factors, which are not directly related to the characteristic of Machine Learning techniques, are as below.

### H. Choice of test and training data

Various reported applications of ML used the following three different means to choose test and training data: *re-substitution* − the same data set is used for both training and testing; *data splitting* − the data set is split into a training set and a test set; independent validation − the model is fitted with a data set independent of the test data set. Naturally, independent validation is the preferred method in most cases, followed by data splitting and then re-substitution. The results obtained by data splitting and re-substitution may be overly optimistic due to over-fitting (Jeschke and Strayer, 2008). However, the choice of one technique over the other is also problem dependent. Only a small segment of the reported studies seems to use independent validation.

### I. Model evaluation metrics

The measure of model performance or the model evaluation technique should ideally be chosen based on the purpose of the study or the modeling exercise. It is thus perfectly understandable that different authors have used different evaluation metric for their specific studies. Pease see the following literature for further discussions on choice of evaluation metrics: Fielding & Bell (1997); Guisan & Zimmermann (2000); Pearce & Ferrier (2000); Manel et al. (2001); Fielding (2002); Liu et al. (2005); Vaughan &

TABLE II.        FACTORS INFLUENCING APPLICATION OF ML TECHNIQUES TO ECOLOGICAL MODELLING

| Factor | Impact on ML technique | Possible solution |
|---|---|---|
| *Very large data sets* | EC, ANN and DT all are adversely effected | Enhanced algorithms, effective sampling, approximation; massively parallel processing |
| *High dimensionality* | EC, ANN and DT all are adversely effected | Reducing dimensionality; using prior knowledge to identify irrelevant attributes |
| *Over-fitting* | DT and some of the ANNs are adversely effected | Cross-validation; regularization |
| *Dynamic environment* | EC, ANN and DT all are adversely effected | Incremental methods capable of updating the patterns and identifying the patterns of changes |
| *Noisy and missing data* | DT is better equipped to handle this problem compared to others | Data smoothing; Statistical strategies to identify hidden variables and dependencies |
| *Complex dependencies among attributes* | EC, ANN and DT all are effected; however, handles better than traditional techniques such as GLM | |
| *Interpretability of the generated patterns* | EC = poor interpretability; DT and ANN= moderate to high interpretability | |
| *Choice of test and training data* | Effects EC, ANN and DT | Depends on goal of the study; however, generally independent validation is better than others |
| *Model evaluation metrics* | Effects EC, ANN and DT | Depends on goal of the study |

Ormerod (2005); Allouche et al. (2006).

Table 2 summarises the factors influencing application of ML techniques to ecological modelling. Also see Jeschke et. al. (Jeschke and Strayer, 2008) for a list on comparative performances of ecological modelling techniques as observed in some of the selected studies found in the literature.

As can be seen, none of the modeling techniques is universally superior compared to other techniques across all applications. Comparative performances of the three traditional methods, namely, GLM, GAM and climate envelope model, shows GAM and GLM have comparable performances. Among the Machine Learning methods, the popular GARP technique produces moderate performance, while CART and ANN have shown mixed results. It may be noted that these examples did not include adequate number of applications of ANN. Jeschke and Strayer (2008) have reported, overall, ANN performs better among the ML techniques applied to this problem domain. Robustness is a characteristic often attributed to ANN. The findings by Jeschke and Strayer (2008) also validate this claim. The specialized climate envelope techniques such as BIOCLIM, FEM and DOMAIN show only moderate performance in general and often perform worse than the Machine Learning techniques. However, some of the relatively recent comparisons have claimed that newer techniques are likely to outperform more established techniques (e.g. the model-averaging random forests by Lawler et al. (2006) and Broennimann et al. (2007); the Bayesian weights-of-evidence model by Zeman & Lynen (2006)). However, as these methods have been used only in a handful of studies, claims about their predictive power is premature (Jeschke and Strayer, 2008). Finally, this comparative study reiterates the fact that success and failure of inductive, data-driven techniques such as the machine learning techniques are primarily dependent on the application, including the complexity and representativeness of the data set and the goal of study.

## IV. CONCLUSION

This paper presented a comprehensive review of applications of various Machine Learning techniques to bioclimatic modelling and broadly to ecological modelling. Some of the statistical techniques popular in this application domain have also been discussed. Factors influencing the performance of such techniques have been identified. It has been concluded that success or failure of application of the Machine Learning techniques to ecological modeling is primarily application dependent and none of techniques can claim superior performance as against other techniques universally. However, the identified factors or characteristics can be used as a guideline to select the ML techniques for such modeling exercises.

Some of the issues that future researches may consider are as follows:

- Hybrid ML techniques have been successfully tried in various applications; this is still underutilized in bioclimatic modelling. Suitable hybrid methods may be useful in handling complexities such as the extreme variability, intermittent and long range correlation involved with the hydro-meteorological fields.

- Goal of the research should be a key driver influencing the choice of the ML technique. For example: ANN would be a good choice where visualization is important; ANN also works well where the intent is to reveal the nature of relationships between the input (driver) and the output variables in the ecosystem; Adaptive agents can be used to predict the structure and behavior of emergent ecosystems in response to environmental changes.

- Machine learning techniques are essentially data-driven techniques. It is important that the dataset is representative of the problem. This includes both the variables considered and the source of data. For example, modelling of species distribution may sometime require pooling of data from populations with very different demographic and ecological history.

- Further research is required about transparency of the modelling process and more importantly the interpretability of the models for ML–based bioclimatic modelling.

Finally, in the bio-climatic modelling context, it is important to remember that the ML techniques are not meant to replace the human experts, but to provide them with powerful tools for prediction, explanation and interpretation of bio-climatic phenomena.